\documentclass[journal]{IEEEtran}
\ifCLASSINFOpdf
\else
\fi
\usepackage{algorithm}
\usepackage{algorithmic}

\usepackage{graphicx}

\begin{document}
%
\title{DLAU: A Scalable Deep Learning Accelerator Unit on FPGA}
%
%
%

\author{Chao~Wang,~\IEEEmembership{Member,~IEEE,}
	        Qi~Yu,~\IEEEmembership{}
		Lei~Gong,~\IEEEmembership{}
        Xi~Li,~\IEEEmembership{Member,~IEEE}
        Yuan~Xie,~\IEEEmembership{Fellow,~IEEE}
        and~Xuehai~Zhou,~\IEEEmembership{Member,~IEEE}
\thanks{C. Wang, Q. Yu, L.Gong, X.Li and X.Zhou are with University of Science and Technology of China, Hefei, 230027, Anhui, China. E-mail: \{cswang,llxx,xhzhou\}@ustc.edu.cn, yuiq@mail.ustc.edu.cn).}
\thanks{Y. Xie is with University of California at Santa Barbara, 93106, United States, E-mail:yuanxie@ece.ucsb.edu.}
\thanks{Manuscript received January 10, 2016.}}

%
%

\markboth{	IEEE Transactions on 
	COMPUTER-AIDED DESIGN 
	of Integrated Circuits and Systems,~Vol.~xx, No.~x, xxxx~2016}%
{Shell \MakeLowercase{\textit{et al.}}: Bare Demo of IEEEtran.cls for IEEE Journals}
%



\maketitle

\begin{abstract}
As the emerging field of machine learning, deep learning shows excellent ability in solving complex learning problems. However, the size of the networks becomes increasingly large scale due to the demands of the practical applications, which poses significant challenge to construct a high performance implementations of deep learning neural networks. In order to improve the performance as well to maintain the low power cost, in this paper we design DLAU, which is a scalable accelerator architecture for large-scale deep learning networks using FPGA as the hardware prototype. The DLAU accelerator employs three pipelined processing units to improve the throughput and utilizes tile techniques to explore locality for deep learning applications. Experimental results on the state-of-the-art Xilinx FPGA board demonstrate that the DLAU accelerator is able to achieve up to 36.1x speedup comparing to the Intel Core2 processors, with the power consumption at 234mW.
\end{abstract}


%
\IEEEpeerreviewmaketitle

\section{Introduction}
%
%
%
%
\IEEEPARstart{I}{n} the past few years, machine learning has become pervasive in various research fields and commercial applications, and achieved satisfactory products. The emergence of deep learning speeded up the development of machine learning and artificial intelligence. Consequently, deep learning has become a research hot spot in research organizations \cite{nature}. In general, deep learning uses a multi-layer neural network model to extract high-level features which are a combination of low-level abstractions to find the distributed data features, in order to solve complex problems in machine learning. Currently the most widely used neural models of deep learning are Deep Neural Networks (DNNs) \cite{service} and Convolution Neural Networks (CNNs) \cite{fpga15}, which have been proved to have excellent capability in solving picture recognition, voice recognition and other complex machine learning tasks.

However, with the increasing accuracy requirements and complexity for the practical applications, the size of the neural networks becomes explosively large scale, such as the Baidu Brain with 100 Billion neuronal connections, and the Google cat-recognizing system with 1 Billion neuronal connections. The explosive volume of data makes the data centers quite power consuming. In particular, the electricity consumption of data centers in U.S. are projected to increase to roughly 140 billion kilowatt-hours annually by 2020 \cite{datacenter}. Therefore, it poses significant challenges to implement high performance deep learning networks with low power cost, especially for large-scale deep learning neural network models.
So far, the state-of-the-art means for accelerating deep learning algorithms are Field-Programmable Gate Array (FPGA), Application Specific Integrated Circuit (ASIC), and Graphic Processing Unit (GPU). Compared with GPU acceleration, hardware accelerators like FPGA and ASIC can achieve at least moderate performance with lower power consumption. However, both FPGA and ASIC have relatively limited computing resources, memory, and I/O bandwidths, therefore it is challenging to develop complex and massive deep neural networks using hardware accelerators. For ASIC, it has a longer development cycle and the flexibility is not satisfying. Chen et al presents a ubiquitous machine-learning hardware accelerator called DianNao \cite{diannao}, which opens a new paradigm to machine learning hardware accelerators focusing on neural networks. But DianNao is not implemented using reconfigurable hardware like FPGA, therefore it cannot adapt to different application demands. Currently around FPGA acceleration researches, Ly and Chow \cite{fpga09} designed FPGA based solutions to accelerate the Restricted Boltzmann Machine (RBM). They created dedicated hardware processing cores which are optimized for the RBM algorithm. Similarly Kim et al \cite{fpl09} also developed a FPGA based accelerator for the restricted Boltzmann machine. They use multiple RBM processing modules in parallel, with each module responsible for a relatively small number of nodes. Other similar works also present FPGA based neural network accelerators \cite{fpga161,fpga162}. Qi et al. present a FPGA based accelerator \cite{ccgrid15}, but it cannot accommodate changing network size and network topologies. To sum up, these studies focus on implementing a particular deep learning algorithm efficiently, but how to increase the size of the neural networks with scalable and flexible hardware architecture has not been properly solved.

To tackle these problems, we present a scalable deep learning accelerator unit named DLAU to speed up the kernel computational parts of deep learning algorithms. In particular, we utilize the tile techniques, FIFO buffers, and pipelines to minimize memory transfer operations, and reuse the computing units to implement the large-size neural networks. This approach distinguishes itself from previous literatures with following contributions:

1.	In order to explore the locality of the deep learning application, we employ tile techniques to partition the large scale input data. The DLAU architecture can be configured to operate different sizes of tile data to leverage the trade-offs between speedup and hardware costs. Consequently the FPGA based accelerator is more scalable to accommodate different machine learning applications.

2.	The DLAU accelerator is composed of three fully pipelined processing units, including TMMU, PSAU, and AFAU. Different network topologies such as CNN, DNN, or even emerging neural networks can be composed from these basic modules. Consequently the scalability of FPGA based accelerator is higher than ASIC based accelerator.

\section{Tile Techniques and Hot Spot Profiling}

Restricted Boltzmann Machines (RBMs) have been widely used to efficiently train each layer of a deep network. Normally a deep neural network is composed of one input layer, several hidden layers and one classifier layer. The units in adjacent layers are all-to-all weighted connected. The prediction process contains feedforward computation from given input neurons to the output neurons with the current network configurations. Training process includes pre-training which locally tune the connection weights between the units in adjacent layers, and global training which globally tune the connection weights with Back Propagation process.

The large-scale deep neural networks include iterative computations which have few conditional branch operations, therefore they are suitable for parallel optimization in hardware. In this paper we first explore the hot spot using the profiler. Results in Fig. \ref{hotspot} illustrates the percentage of running time including Matrix Multiplication (MM), Activation, and Vector operations. For the representative three key operations: feed forward, Restricted Boltzmann Machine (RBM), and back propagation (BP), matrix multiplication play a significant role of the overall execution. In particular, it takes 98.6\%, 98.2\%, and 99.1\% of the feed forward, RBM, and BP operations. In comparison, the activation function only takes 1.40\%, 1.48\%, and 0.42\% of the three operations. Experimental results on profiling demonstrate that the design and implementation of MM accelerators is able to improve the overall speedup of the system significantly.

\begin{table}[!t]
	\renewcommand{\arraystretch}{1.3}
	\caption{Profiling of Hot Spots of DNN}
	\label{hotspot}
	\centering
	\begin{tabular}{|c||c||c||c|}
		\hline
			Algorithms & Matrix Multiplication &	Activation &	Vector\\ \hline
			Feedforward	&		98.60\% &		1.40\% &	\\ \hline
			RBM	&		98.20\% &		1.48\% &		0.30\% \\ \hline
			BP	&		99.10\% &		0.42\% &		0.48\% \\ \hline
			
	\end{tabular}
\end{table}

However, considerable memory bandwidth and computing resources are needed to support the parallel processing, consequently it poses a significant challenge to FPGA implementations compared with GPU and CPU optimization measures. In order to tackle the problem, in this paper we employ tile techniques to partition the massive input data set into tiled subsets. Each designed hardware accelerator is able to buffer the tiled subset of data for processing. In order to support the large-scale neural networks, the accelerator architecture are reused. Moreover, the data access for each tiled subset can run in parallel to the computation of the hardware accelerators.

\begin{algorithm}
	\caption{Pseudocode Code of the Tiled Inputs}
	\begin{algorithmic} 
		\REQUIRE\
		
		Ni: the number of the input neurons
		
		No: the number of the output neurons
		
		Tile\_Size: the tile size of the input data
		
		batchsize: the batch size of the input data
		
		\FOR {$n=0;n<batchsize;n++$}
		\FOR {$k=0;k<Ni;k+=Tile\_Size$} 
		\FOR {$j=0;j<No;j++$} 
		\STATE $	   y[n][j]=0;$ \\
		\FOR {$i=k;i<k+Tile\_Size\&\&i<Ni;i++$} 
		\STATE	$		     y[n][j]+=w[i][j]*x[n][i]  $
		\IF{$i==Ni-1$ }	
		\STATE	$                            y[n][j]=f(y[n][j]);$
		\ENDIF
		\ENDFOR
		\ENDFOR
		\ENDFOR	
		\ENDFOR

	\end{algorithmic}
\end{algorithm}

In particular, for each iteration, output neurons are reused as the input neurons in next iteration. To generate the output neurons for each iteration, we need to multiply the input neurons by each column in weights matrix. As illustrated in Algorithm 1, the input data are partitioned into tiles and then multiplied by the corresponding weights. Thereafter the calculated part sum are accumulated to get the result. Besides the input/output neurons, we also divided the weight matrix into tiles corresponding to the tile size. As a consequence, the hardware cost of the accelerator only depends on the tile size, which saves significant number of hardware resources. The tiled technique is able to solve the problem by implementing large networks with limited hardware. Moreover, the pipelined hardware implementation is another advantage of FPGA technology compared to GPU architecture, which uses massive parallel SIMD architectures to improve the overall performance and throughput. According to the profiling results depicted in Table \ref{hotspot}, during the prediction process and the training process in deep learning algorithms, the common but important computational parts are matrix multiplication and activation functions, consequently in this paper we implement the specialized accelerator to speed up the matrix multiplication and activation functions.

\section{DLAU Architecture and Execution Model}

\begin{figure}[!t]
	\centering
	\includegraphics[width=3in]{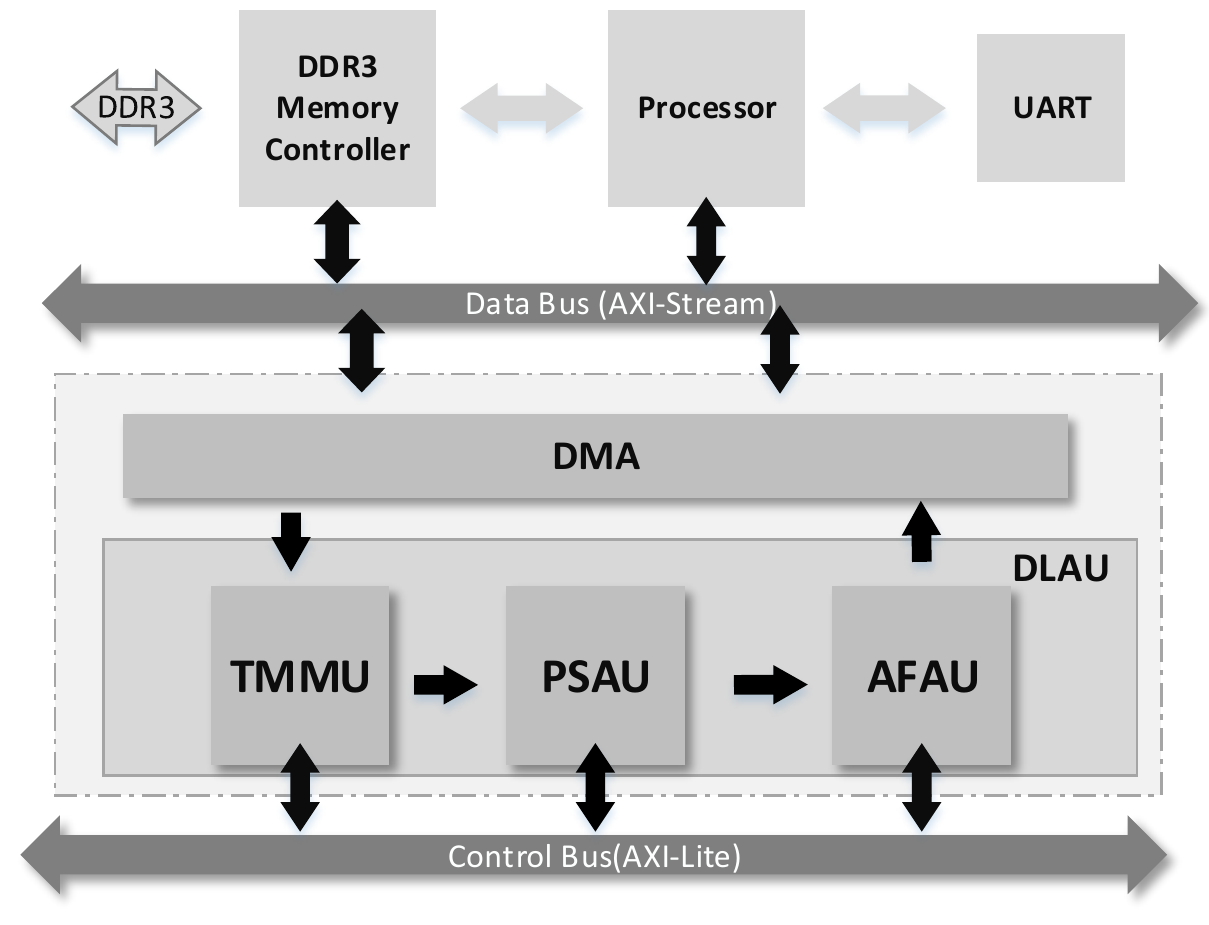}
	\caption{DLAU Accelerator Architecture.}
	\label{dlau}
\end{figure}

Fig. \ref{dlau} describes the DLAU system architecture which contains an embedded processor, a DDR3 memory controller, a DMA module, and the DLAU accelerator. The embedded processor is responsible for providing programming interface to the users and communicating with DLAU via JTAG-UART. In particular it transfers the input data and the weight matrix to internal BRAM blocks, activates the DLAU accelerator, and returns the results to the user after execution. The DLAU is integrated as a standalone unit which is flexible and adaptive to accommodate different applications with configurations. The DLAU consists of 3 processing units organized in a pipeline manner: Tiled Matrix Multiplication Unit (TMMU), Part Sum Accumulation Unit (PSAU), and Activation Function Acceleration Unit (AFAU). For execution, DLAU reads the tiled data from the memory by DMA, computes with all the three processing units in turn, and then writes the results back to the memory.

In particular, the DLAU accelerator architecture has following key features:

\textbf{FIFO Buffer: }Each processing unit in DLAU has an input buffer and an output buffer to receive or send the data in FIFO. These buffers are employed to prevent the data loss caused by the inconsistent throughput between each processing unit.

\textbf{Tiled Techniques: }Different machine learning applications may require specific neural net-work sizes. The tile technique is employed to divide the large volume of data into small tiles that can be cached on chip, therefore the accelerator can be adopted to different neural network size. Consequently the FPGA based accelerator is more scalable to accommodate different machine learning applications.

\textbf{Pipeline Accelerator: }We use stream-like data passing mechanism (e.g. AXI-Stream for demonstration) to transfer data between the adjacent processing units, therefore TMMU, PSAU, and AFAU can compute in streaming-like manner. Of these three computational modules, TMMU is the primary computational unit, which reads the total weights and tiled nodes data through DMA, performs the calculations, and then transfers the intermediate Part Sum results to PSAU. PSAU collects Part Sums and performs accumulation. When the accumulation is completed, results will be passed to AFAU. AFAU performs the activation function using piecewise linear interpolation methods. In the rest of this section, we will detail the implementation of these three processing units respectively.

\subsection{TMMU architecture}
\begin{figure}[!t]
	\centering
	\includegraphics[width=3in]{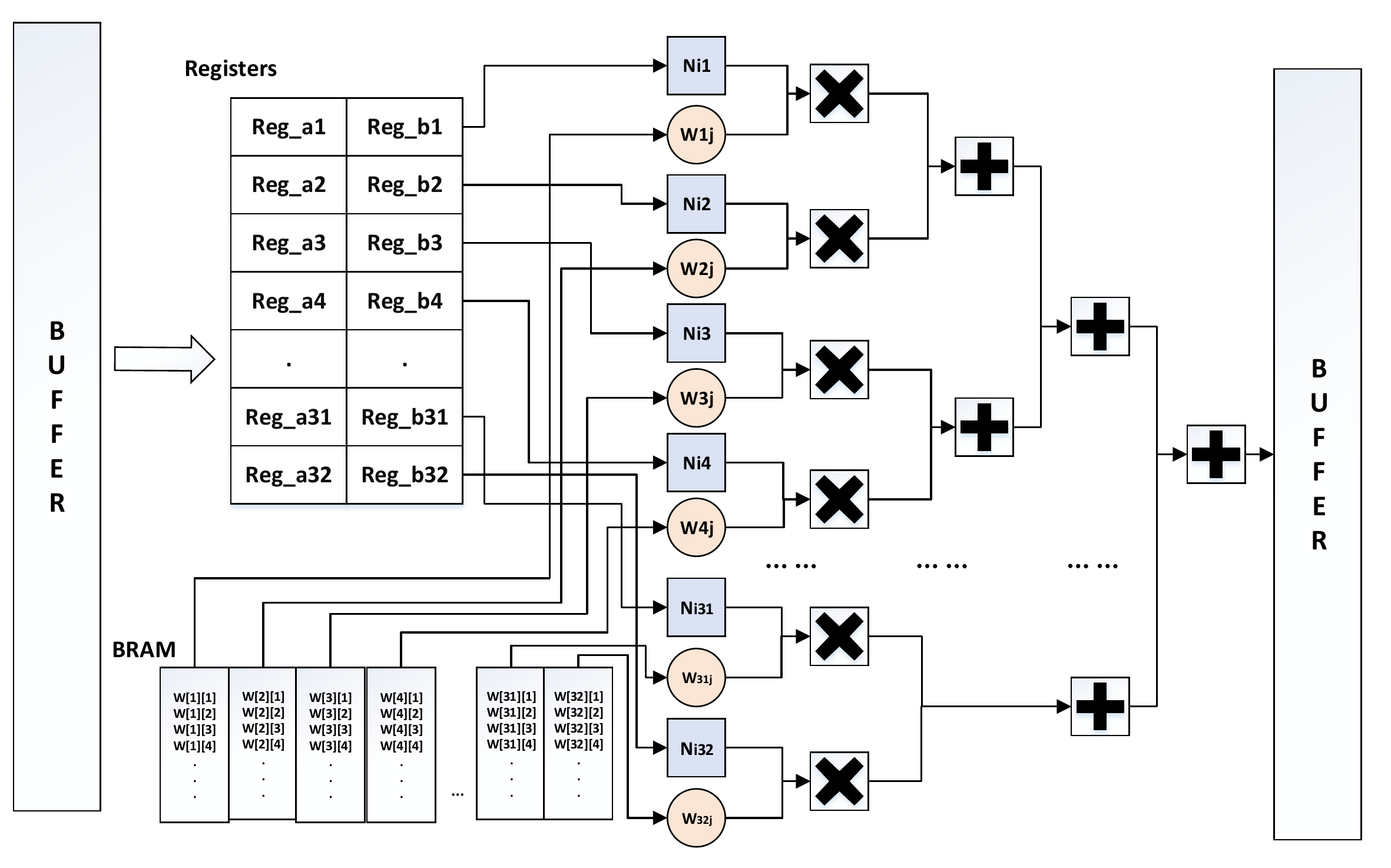}
	\caption{TMMU Schematic Diagram.}
	\label{tmmu}
\end{figure}

Tiled Matrix Multiplication Unit (TMMU) is in charge of multiplication and accumulation operations. TMMU is specially designed to exploit the data locality of the weights and is responsible for calculating the Part Sums. TMMU employs an input FIFO buffer which receives the data transferred from DMA and an output FIFO buffer to send Part Sums to PSAU. Fig. \ref{tmmu} illustrates the TMMU schematic diagram, in which we set tile size=32 as an example. TMMU firstly reads the weight matrix data from input buffer into different BRAMs in 32 by the row number of the weight matrix (n=i\%32，where n refers to the number of BRAM, and i is the row number of weight matrix). Then, TMMU begins to buffer the tiled node data. In the first time, TMMU reads the tiled 32 values to registers Reg\_a and starts execution. In parallel to the computation at every cycle, TMMU reads the next node from input buffer and saves to the registers Reg\_b. Consequently the registers Reg\_a and Reg\_b can be used alternately.

For the calculation, we use pipelined binary adder tree structure to optimize the performance. As depicted in Fig. \ref{tmmu}, the weight data and the node data are saved in BRAMs and registers. The pipeline takes advantage of time-sharing the coarse-grained accelerators. As a consequence, this implementation enables the TMMU unit to produce a Part Sum result every clock cycle.

\subsection{PSAU architecture}
\begin{figure}[!t]
	\centering
	\includegraphics[width=2.5in]{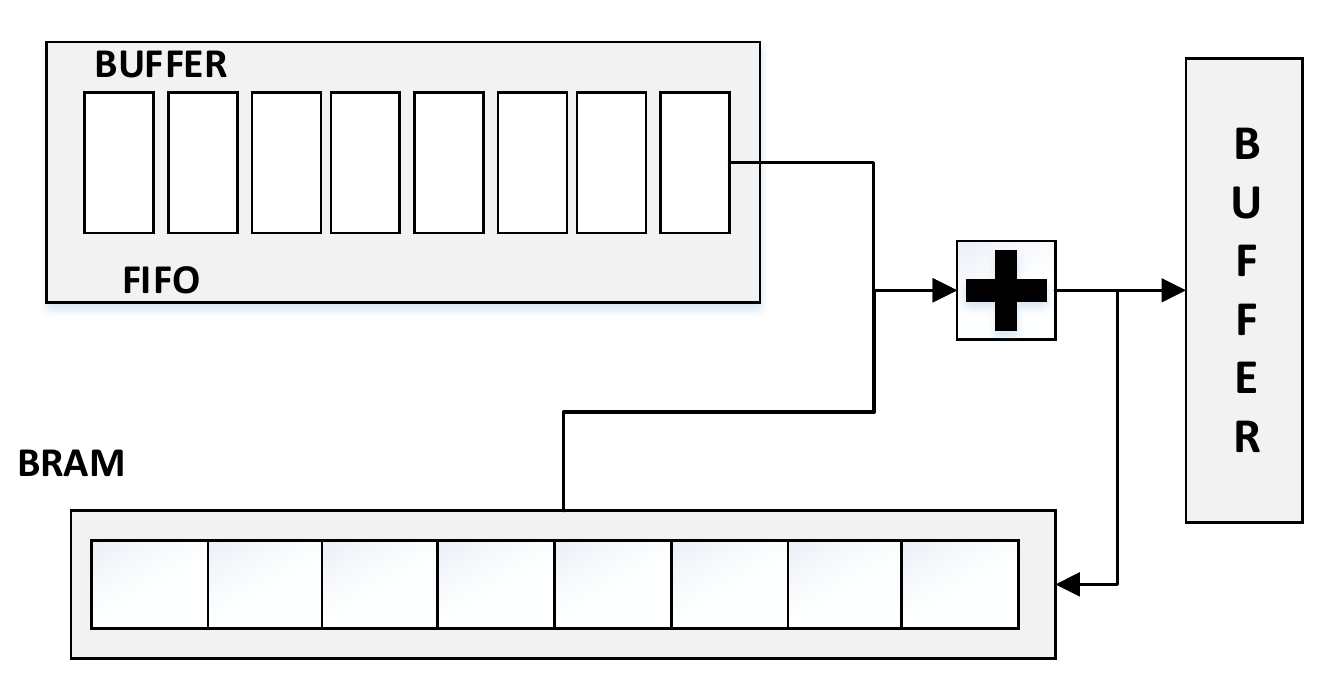}
	\caption{PSAU Schematic Diagram}
	\label{psau}
\end{figure}

Part Sum Accumulation Unit (PSAU) is responsible for the accumulation operation. Fig. \ref{psau} presents the PSAU architecture, which accumulates the part sum produced by TMMU. If the Part Sum is the final result, PSAU will write the value to output buffer and send results to AFAU in a pipeline manner. PSAU can accumulate one Part Sum every clock cycle, therefore the throughput of PSAU accumulation matches the generation of the Part Sum in TMMU.

\subsection{AFAU architecture}

Finally, Activation Function Acceleration Unit (AFAU) implements the activation function using piecewise linear interpolation (y=ai*x+bi, x$\in$[x$_{1}$,x$_{i+1}$)). This method has been widely applied to implement activation functions with negligible accuracy loss when the interval between x$_{i}$ and x$_{i+1}$ is insignificant. Eq. (1) shows the implementation of sigmoid function. For x$>$8 and x$\le$-8, the results are sufficiently close to the bounds of 1 and 0, respectively. For the cases in -8$<$x$\le$0 and 0$<$x$\le$8, different functions are configured. In total we divide the sigmoid function into four segments.

\begin{equation}
f(x)=
\left\{
\begin{array}{r@{\;~~\;}l}
0 &\mbox{if $x\le-8$}\\
1+a$[$\lfloor\frac{-x}{k} \rfloor$]$x-b$[$\lfloor\frac{-x}{k}\rfloor$]$&\mbox{if $-8<x\le0$}\\
a$[$\lfloor\frac{x}{k}\rfloor $]$x+b$[$\lfloor\frac{x}{k}\rfloor$]$&\mbox{if $0<x\le8$}\\
1 &\mbox{if $x>8$}\\
\end{array}
\right.
\end{equation}

Similar to PSAU, AFAU also has both input buffer and output buffer to maintain the throughput with other processing units. In particular, we use two separate BRAMs to store the values of a and b. The computation of AFAU is pipelined to operate sigmoid function every clock cycle. As a consequence, all the three processing units are fully pipelined to ensure the peak throughput of the DLAU accelerator architecture.

\section{Experiments and Data Analysis}

In order to evaluate the performance and cost of the DLAU accelerator, we have implemented the hardware prototype on the Xilinx Zynq Zedboard development board, which equips ARM Cortex-A9 processors clocked at 667MHz and programmable fabrics. For benchmarks, we use the Mnist data set to train the 784$\times$M$\times$N$\times$10 Deep Neural Networks in Matlab, and use M$\times$N layers’ weights and nodes value for the input data of DLAU. For comparison, we use Intel Core2 processor clocked at 2.3GHz as the baseline.

In the experiment we use Tile size=32 considering the hardware resources integrated in the Zedboard development board. The DLAU computes 32 hardware neurons with 32 weights every cycle. The clock of DLAU is 200MHz (one cycle takes 5ns). Three network sizes---64$\times$64, 128$\times$128, and 256$\times$256 are tested. 

\subsection{Speedup Analysis}
We present the speedup of DLAU and some other similar implementations of the deep learning algorithms in Table \ref{comparison}. Experimental results demonstrate that the DLAU is able to achieve up to 36.1x speedup at 256$\times$256 network size. In comparison, Ly\&Chow’s work \cite{fpga09} and Kim et.al’s work \cite{fpl09} present the work only on Restricted Boltzmann Machine algorithms, while the DLAU is much more scalable and flexible. DianNao \cite{diannao} reaches up to 117.87x speedup due to its high working frequency at 0.98GHz. Moreover, as DianNao is hardwired instead of implemented on a FPGA platform, therefore it cannot efficiently adapt to different neural network sizes.

\begin{table}[!t]
	\renewcommand{\arraystretch}{1.3}
	\caption{Comparisons between Similar Approaches}
	\label{comparison}
	\centering
	\begin{tabular}{|c||c||c||c||c|}
		\hline
		\textbf{Work} & \textbf{Network}  & \textbf{Clock} 	& \textbf{Speedup} & \textbf{Baseline}\\
		\hline
		Ly\&Chow \cite{fpga09} & 256$\times$256 & 100MHz	& 32$\times$	& 2.8GHz P4\\
		\hline
		Kim et.al \cite{fpl09} & 256$\times$256	& 200MHz&	25$\times$&	2.4GHz Core2\\
		\hline
		DianNao \cite{diannao}& General&	0.98GHz&	117.87$\times$&	2GHz SIMD\\
		\hline
		Zhang et.al \cite{fpga15}& 256$\times$256&	100MHz&	17.42$\times$&	2.2GHz Xeon\\
		\hline
		DLAU&	256$\times$256&	200MHz&	36.1$\times$&	2.3GHz Core2\\
		\hline
	\end{tabular}
\end{table}

\begin{figure}[!t]
	\centering
	\includegraphics[width=3.3in]{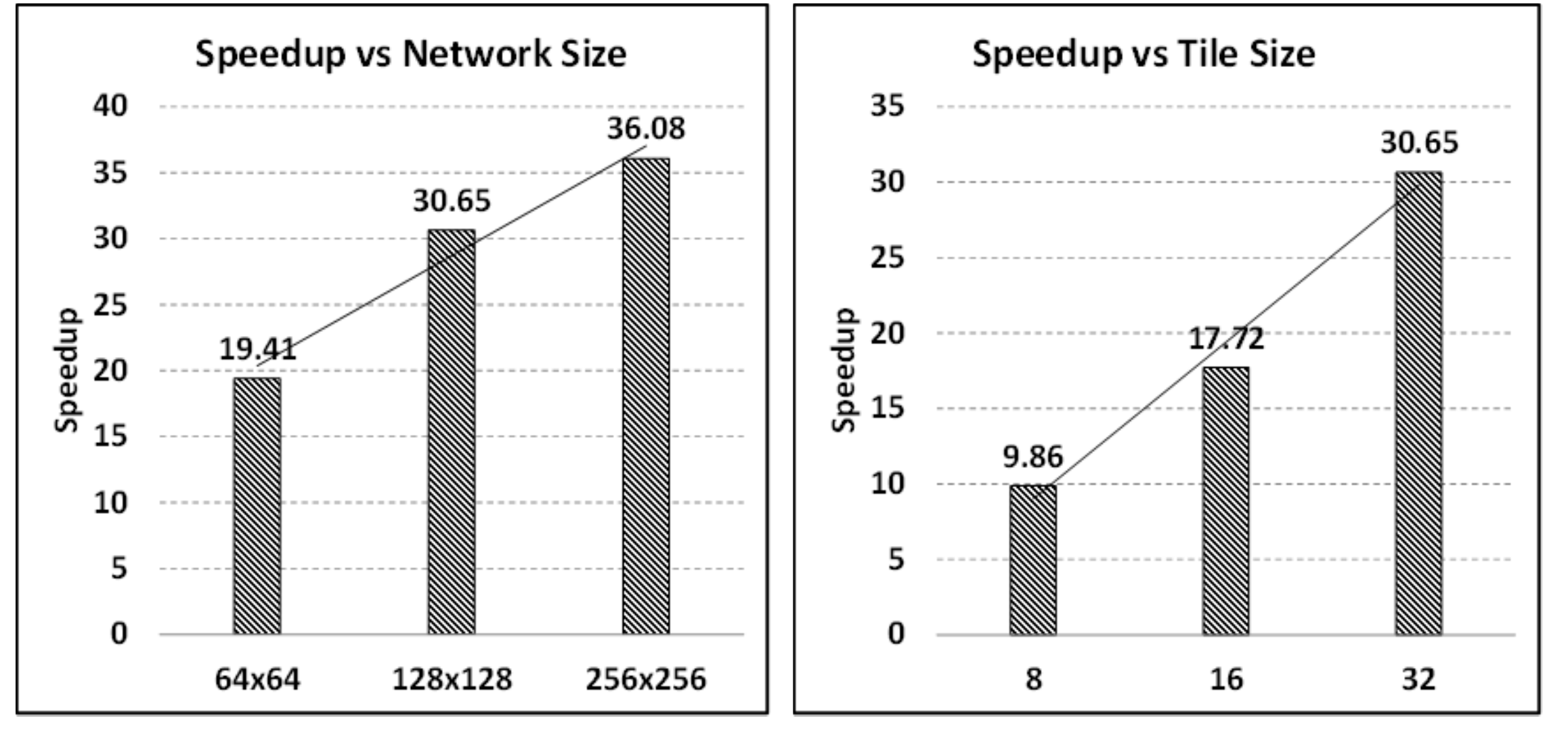}
	\caption{Speedup at Different Network Sizes and Tile Sizes.}
	\label{speedup}
\end{figure}

Fig. \ref{speedup} illustrates the speedup of DLAU at different network sizes-64$\times$64, 128$\times$128, and 256$\times$256 respectively. Experimental results demonstrate a reasonable ascendant speedup with the growth of neural networks sizes. In particular, the speedup increases from 19.2x in 64$\times$64 network size to 36.1x at the 256$\times$256 network size. The right part of Fig. \ref{speedup} illustrates how the tile size has an impact on the performance of the DLAU. It can be acknowledged that bigger tile size means more number of neurons to be computed concurrently. At the network size of 128$\times$128, the speedup is 9.2x when the tile size is 8. When the tile size increases to 32, the speedup reaches 30.5x. Experimental results demonstrate that the DLAU framework is configurable and scalable with different tile sizes. The speedup can be leveraged with hardware cost to achieve satisfying trade-offs.

\subsection{Resource utilization and Power}

Table \ref{resource} summarizes the resource utilization of DLAU in 32$\times$32 tile size including the BRAM resources, DSPs, FFs, and LUTs. TMMU is much more complex than the rest two hardware modules therefore it consumes most hardware resources. Taking the limited number of hardware logic resources provided by Xilinx XC7Z020 FPGA chip, the overall utilization is reasonable. The DLAU utilizes 167 DSP blocks due to the use of the Floating-point addition and the Floating-point multiplication operations. 

\begin{table}[!t]
	\renewcommand{\arraystretch}{1.3}
	\caption{Resource Utilization of DLAU at 32$\times$32 Tile Size}
	\label{resource}
	\centering
	\begin{tabular}{|c||c||c||c||c|}
		\hline
		\textbf{Component}&	\textbf{BRAMs}&	\textbf{DSPs}&	\textbf{FFs}&	\textbf{LUTs}\\
		\hline
		TMMU&	32&	158&	25356&	32461\\
		\hline
		PSAU&	1&	2&	754&	632\\
		\hline
		AFAU&	2&	7&	2216&	3291\\
		\hline
		Total&	35&	167&	28326&	36384\\
		\hline
		Available&	280&	220&	106400&	53200\\
		\hline
		Utilization&	12.5\%&	75.9\%&	26.6\%&	68.4\%\\
		\hline
	\end{tabular}
\end{table}

Table \ref{resourcecom} compares the resource utilization of DLAU with other two FPGA based literatures. Experimental results depict that our DLAU accelerator occupies similar number of FFs and LUTs to Ly\&Chow's work \cite{fpga09}, while it only consumes 35/257=13.6\% on the BRAMs. Comparing to the Kim et.al's work \cite{fpl09}, the BRAM utilization of DLAU is insignificant. This is due to the tile techniques so that large scale neural networks can be divided into small tiles, therefore the scalability and flexibility of the architecture is significantly improved.

\begin{table}[!t]
	\renewcommand{\arraystretch}{1.3}
	\caption{Resource Comparisons between Similar Approaches}
	\label{resourcecom}
	\centering
	\begin{tabular}{|c||c||c||c||c||c|}
		\hline
		\textbf{Implementation}& \textbf{FPGA}&	\textbf{BRAMs}&	\textbf{DSPs}&	\textbf{FFs}&	\textbf{LUTs}\\
		\hline
		Ly\&Chow [5] & XC2VP70&	257&	N/A&	30403&	29885\\
		\hline
		Kim et.al [7] & N/A&	589824&	18&	11790&	7662\\
		\hline
		DLAU&	XC7Z020&	35&	167&	28326&	36384\\
		\hline
	\end{tabular}
\end{table}


\begin{table}[!t]
	\renewcommand{\arraystretch}{1.3}
	\label{totalpower}
	\caption{Power Consumption of the Units}
	\centering
	\begin{tabular}{|c|c||c|c|}
		\hline
		\textbf{Component}&	\textbf{Power}	&\textbf{Component}&\textbf{Power}\\
		\hline
		Accelerator-TMMU&	189mW	&	Processor&	1307mW\\
		\hline
		Accelerator-PSAU&	5mW&	DDR Controller&	177mW\\
		\hline
		Accelerator-AFAU&	25mW	&	Peripherals&	26mW\\
		\hline
		Accelerator-DMA&	15mW	&	Clocks&	70mW\\
		\hline
		Accelerator-Total&	234mW	&	System Total&	1814mW\\
		\hline
		
	\end{tabular}
\end{table}

\begin{figure}[!t]
	\centering
	\includegraphics[width=3.5in]{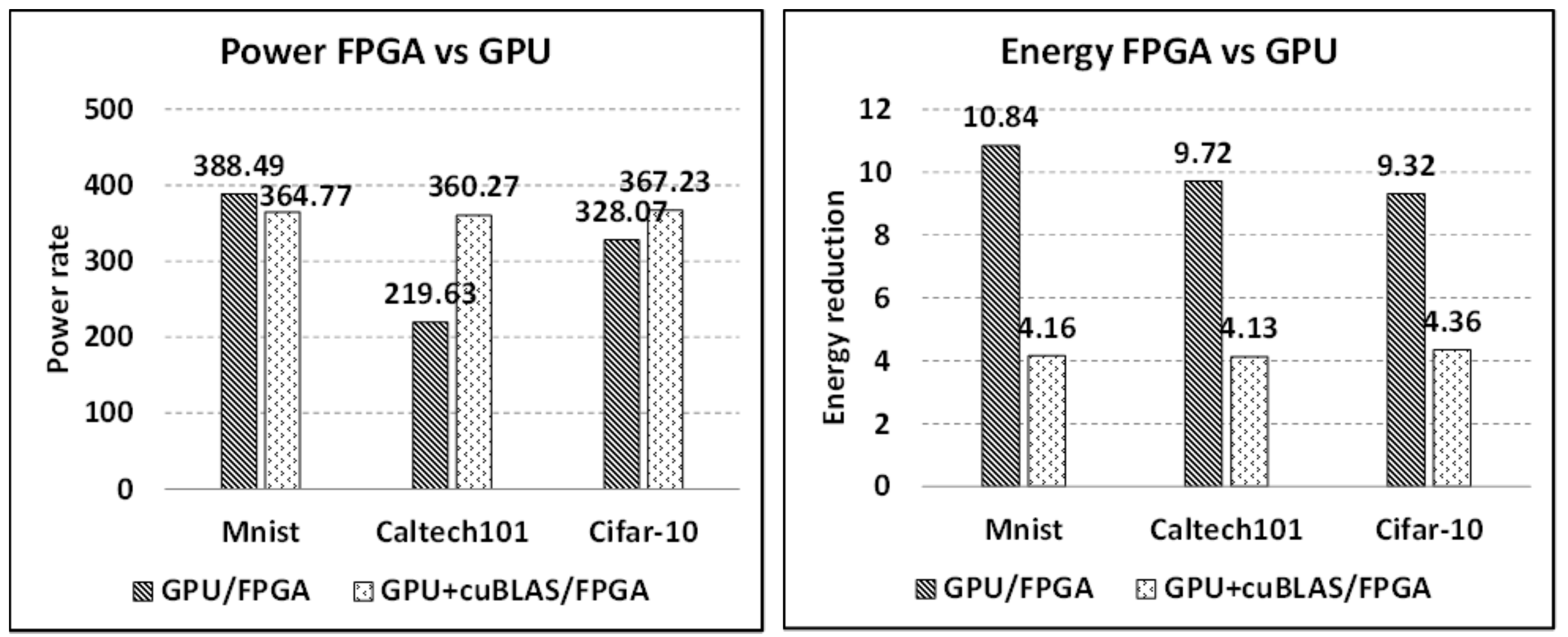}
	\caption{Power and Energy Comparison between FPGA and GPU }
	\label{energy}
\end{figure}

In order to evaluate the power consumption of accelerator, we use Xilinx Vivado tool set to achieve power cost of each processing unit in DLAU and the DMA module. The results in Table \ref{totalpower} depict that the total power of DLAU is only 234mW, which is much lower than that of DianNao (485mW). The results demonstrate that the DLAU is quite energy efficient as well as highly scalable compared to other accelerating techniques. To compare the energy and power between FPGA based accelerator and GPU based accelerators, we also implement a prototype using the state-of-the-art NVIDIA Tesla K40c as the baseline. K40c has 2880 stream cores working at peak frequency 875MHz, and the Max Memory Bandwidth is 288 (GB/sec). In comparison, we only employ 1 DLAU on the FPGA board working at 100MHz. In order to evaluate the speedup of the accelerators in a real deep learning applications, we use DNN to model 3 benchmarks, including Caltech101, Cifar-10, and MNIST, respectively. Fig. \ref{energy} illustrates the comparison between FPGA based GPU+cuBLAS implementations. It reveals that the power consumption of GPU based accelerator is 364 times higher than FPGA based accelerators. Regarding the total energy consumption, the FPGA based accelerator is 10x more energy efficient than GPU, and 4.2x than GPU+cuBLAS optimizations.

\begin{figure}[!t]
	\centering
	\includegraphics[width=3.5in]{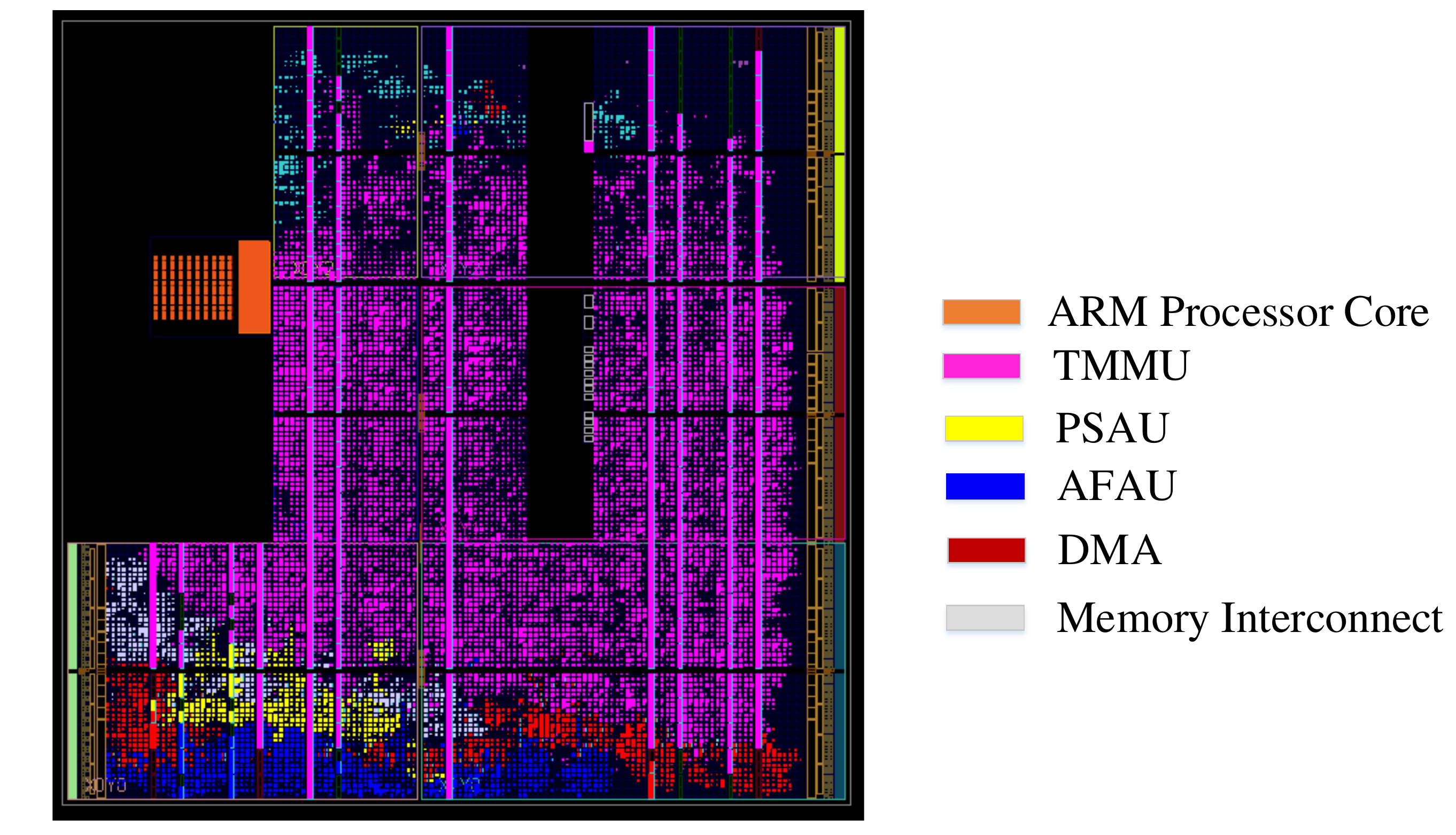}
	\caption{Floorplan of the FPGA Chip}
	\label{floorplan}
\end{figure}

Finally Fig. \ref{floorplan} illustrates the floor plan of the FPGA chip. The left corner depicts the ARM processor which is hardwired in the FPGA chip. Other modules, including different components of the DLAU accelerator, the DMA, and memory interconnect, are presented in different colors. Regarding the programming logic devices, TMMU takes most of the areas as it utilizes a significant number of LUTs and FFs.

\section{Conclusion and Future Work}
In this article we have presented DLAU, which is a scalable and flexible deep learning accelerator based on FPGA. The DLAU includes three pipelined processing units, which can be reused for large scale neural networks. DLAU uses tile techniques to partition the input node data into smaller sets and compute repeatedly by time-sharing the arithmetic logic. Experimental results on Xilinx FPGA prototype show that DLAU can achieve 36.1x speedup with reasonable hardware cost and low power utilization.

The results are promising but there are still some future directions, including optimization of the weight matrix and memory access. Also the trade-off analysis between FPGA and GPU accelerators is another promising direction for large scale neural networks accelerations.

\ifCLASSOPTIONcaptionsoff
  \newpage
\fi

\end{document}